\documentclass{article}

\usepackage{microtype}
\usepackage{graphicx}
\usepackage{subfigure}
\usepackage{multicol}
\usepackage{amsmath}
\usepackage{cuted}
\usepackage{booktabs} 

\usepackage{hyperref}

\usepackage[accepted]{icml2025}

\usepackage{amsmath}
\usepackage{amssymb}
\usepackage{mathtools}
\usepackage{amsthm}
\usepackage{mathtools}

\usepackage[capitalize,noabbrev]{cleveref}

\theoremstyle{plain}

\theoremstyle{definition}

\theoremstyle{remark}

\newcommand{\R}{\mathbb{R}}

\newcommand{\FF}{\textrm{FF}}

\usepackage[textsize=tiny]{todonotes}

\icmltitlerunning{Phase Transitions in Large Language Models and the $O(N)$ Model}

\begin{document}

\twocolumn[
\icmltitle{Phase Transitions in Large Language Models and the $O(N)$ Model}


\begin{icmlauthorlist}
\icmlauthor{Youran Sun}{ymsc,msra}
\icmlauthor{Babak Haghighat}{ymsc,bimsa}
\end{icmlauthorlist}

\icmlaffiliation{ymsc}{Yau Mathematical Sciences Center, Tsinghua University, Beijing, 100084, China}
\icmlaffiliation{bimsa}{Beijing Institute of Mathematical Sciences and Applications (BIMSA), Huairou District, Beijing, 101408, China}
\icmlaffiliation{msra}{Microsoft Research Asia}

\icmlcorrespondingauthor{Babak Haghighat}{babakhaghighat@tsinghua.edu.cn}

\icmlkeywords{Machine Learning, ICML}

\vskip 0.3in
]



\printAffiliationsAndNotice{\icmlEqualContribution} 

\begin{abstract}

Large language models (LLMs) exhibit unprecedentedly rich scaling behaviors.
In physics, scaling behavior is closely related to phase transitions, critical phenomena, and field theory.
To investigate the phase transition phenomena in LLMs, we reformulated the Transformer architecture as an $O(N)$ model.
Our study reveals two distinct phase transitions corresponding to the temperature used in text generation and the model's parameter size, respectively.
The first phase transition enables us to estimate the internal dimension of the model, while the second phase transition is of \textit{higher-depth} and signals the emergence of new capabilities.
As an application, the energy of the $O(N)$ model can be used to evaluate whether an LLM's parameters are sufficient to learn the training data.

\end{abstract}

\section{Introduction}
\label{introduction}

Natural language processing based on Transformer architectures has achieved remarkable success in recent years, leading to numerous industrial applications \cite{google_bert_search_2019, gpt3_2020}. 
The Transformer exhibits unprecedentedly rich scaling law behaviors as an independent research subject.
Scaling laws describe a power-law dependency between two variables, \(y\) and \(x\), near specific points in parameter space, expressed as \(y / y_0 = (x / x_0)^\alpha\), or equivalently as a linear relationship in the logarithmic form \(\log y = \alpha \log x + b\).
In physics, \(\alpha\) is referred to as the \textit{critical exponent}. 
We use Greek letters such as \(\alpha, \beta\), etc., to represent critical exponents.
Some of the most significant scaling laws for large language models (LLMs) discovered to date include
\begin{enumerate}
\itemsep0em 
    \item Test Loss $L$ v.s. Number of Parameters $P$: \(L \sim P^{-\alpha}\), where \(\alpha \approx 0.076\) \cite{scaling_openai_2020} or $0.072$ \cite{scaling_rosenfeld_2021};
    \item Test Loss $L$ v.s. Dataset Size $D$: $L \sim D^{-\beta}$, where $\beta\approx 0.095$ \cite{scaling_openai_2020} or $0.12$ \cite{scaling_rosenfeld_2021};
    \item Test Loss $L$ v.s. Train Compute $C$: $L \sim C^{-\gamma}$, where $\gamma\approx 0.050$ \cite{scaling_openai_2020};
    \item Test Error $\mathcal{E}$ v.s. Inference Compute $C_{i}$: $\mathcal{E}\sim C_{i}^{\delta'}$ where $\delta' \approx 0.164$ \cite{scaling_tsinghua2024} or $0.17$ \cite{deepseekr1_2024}.
\end{enumerate}
We can see that the critical exponents from different sources agree with each other very well.
Combining the results in \cite{scaling_openai_2020, scaling_tsinghua2024}, we can get more critical exponents, for example
\begin{equation}\begin{aligned}
    \mathcal{E}&\sim L^{\epsilon}, \quad \epsilon\approx 2.6;\\
    L&\sim C_{i}^{-\delta},\quad \delta\approx 0.064.
\end{aligned}
\end{equation}
Putting $P$ and $D$ together, \cite{scaling_openai_2020} also proposed the following empirical formula for $L(P,D)$
\begin{equation}
    L(P,D)=\left(\left(\frac{P_c}{P}\right)^{\frac{\alpha}{\beta}}+\frac{D_c}{D}\right)^\beta.
\end{equation}

In physics, critical exponents are almost synonymous with phase transitions.
This concept dates back to 1893 when van der Waals first used critical exponents to characterize the phase transition of van der Waals gas.
In 1895, Pierre Curie studied demagnetization by heat and pointed out the similarity between liquid-gas and ferromagnetic transitions.
In 1937, Landau proposed the mean-field theory of phase transitions.
In 1944, Onsager provided the exact solution to the 2D Ising model, which the \(O(N)\) model studied in this paper generalizes. 
By the 1970s, the concept of the renormalization group (RG) was developed, explaining universal properties of phase transitions, including critical exponents.
In the 1980s, conformal field theory (CFT) was developed, which can be used to classify continuous phase transitions (i.e. universality class).
Readers may refer to \cite{nlab_2019,tao_2010} for more comprehensive historical reviews.  

\begin{figure*}
    \centering
    \vspace{2ex}\hspace{0.09\linewidth}
    \includegraphics[width=0.9\linewidth]{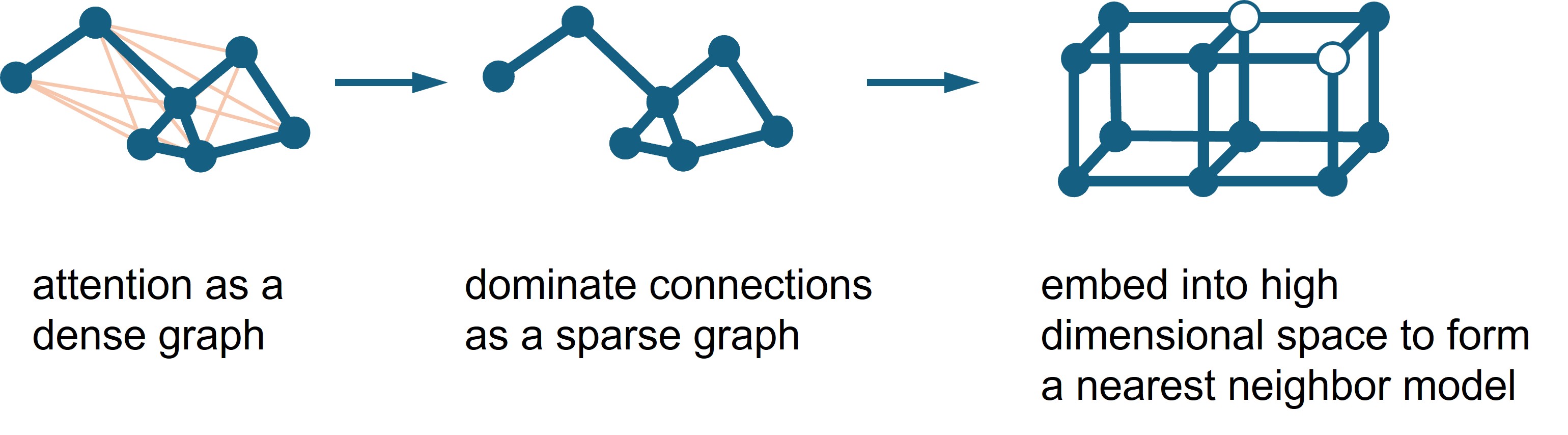}
    \vspace{-2ex} 
    \caption{Demonstartion of how to reformulate Transformer as an $O(N)$ model.}
    \label{fig:transformerasOn}
\end{figure*}

To investigate phase transition phenomena in Transformers, we reformulated the Transformer architecture as an \(O(N)\) model, as demonstrated in Figure \ref{fig:transformerasOn}.
The \(O(N)\) model is a system of interacting spins on a lattice as a generalization of the Ising model.
We can define and measure the energy, susceptibility, and specific heat of Transformers via the $O(N)$ model.

We theoretically and experimentally analyzed the phase transition behaviors of Transformers as \(O(N)\) models, uncovering two distinct phase transitions.
The first phase transition is related to the temperature used during text generation, which enables us to estimate the dimension of the \(O(N)\) model corresponding to the Transformer, referred to as the \textit{internal dimension}.
The second phase transition is linked to the model's parameter size $P$ and represents a phase transition of the phase transition, which we term the \textit{higher-depth phase transition}.
We observed that when the model's parameter size becomes sufficiently large, it exhibits behaviors absent in smaller models --- an emergent phenomenon \cite{emergent_google_2022}.

As a practical application of the theory, the proposed energy can serve as a training indicator.
The \(E\)-\(T\) curve can be obtained within minutes without relying on additional test sets and provides valuable insights into the model's training status.
Furthermore, our experiments reveal that large and small models are fundamentally different.
Although both are based on the Transformer architecture, once the parameter size exceeds a critical threshold \(P_c\approx 7\textrm{B}\), large models exhibit capabilities absent in smaller models.

\paragraph{Contributions.} The main contributions of this paper are
\begin{enumerate}
\itemsep0em 
    \item We reformulate the Transformer as an $O(N)$ model and estimate the internal dimension of the Transformer.
    \item We find a phase transition with respect to the temperature used in generating text. This can be used as a training indicator.
    \item We find a higher-depth phase transition with respect to the number of parameters. This is new evidence of the emergence of capabilities.
\end{enumerate}

The organization of the paper is as follows.
Section \ref{sec:transformer} provides a detailed explanation of the Transformer architecture.  
Section \ref{sec:reviewOn} introduces the \(O(N)\) model.
Together, these two sections aim to make the paper accessible to readers with both physics and computer science backgrounds.  
Section \ref{sec:transformerasOn} explains how the Transformer can be interpreted as an \(O(N)\) model.  
Section \ref{sec:experiments} presents our experimental results.  
Section \ref{sec:relatedwork} reviews theoretical work on LLMs and scaling laws.
Finally, Section \ref{sec:conclusion} provides our conclusions and an outlook for future research.  


\section{Transformer}
\label{sec:transformer}

The Transformer \cite{transformer17} is the standard architecture to deal with natural languages now.
This section will be used to introduce the transformer in great detail.
The pipeline of a Transformer can be roughly divided into tokenization, token embedding, and attention layers.

\paragraph{Tokenization and Token Embedding} For a machine to understand human language, we have to transform a sentence into numbers. Given a sentence $s$, the tokenizer will divide it into many \textit{tokens}.
Simply speaking, a token is a subword.
For example, 
\begin{equation}\begin{aligned}
    \mathtt{dreaming}&\xrightarrow{\text{tokenizer}} \mathtt{dream, ing},\\
    \mathtt{carefully}&\xrightarrow{\text{tokenizer}} \mathtt{care, ful, ly}.
\end{aligned}
\end{equation}
Another way to understand the concept of a token is that its granularity aligns with the "characters" in Chinese.
The number of all the tokens in an LLM is called \textit{vocabulary size}, denoted by $V$.

Then, the token embedding will map each token into a high dimensional space, $\R^N$, where the $N$ is called \textit{embedding dimension}.
Or we can give each token an index between 0 and $V$; then, one can say that the token embedding is a $V\times N$ matrix.
After tokenization and token embedding, a piece of human language is transformed into a list of vectors; each vector is in $\R^N$, and the list length is the number of tokens in the sentence.
For example
\begin{equation}
\begin{aligned}
    s =& \;\mathtt{Hello~World!}\\
    \xrightarrow{\text{tokenizer}} ~& \mathtt{[Hello, world, !] or [7592, 2088,  999]} \\
    \xrightarrow{\text{token embedding}} ~& \textrm{a tensor of shape } (3, N)
\end{aligned}
\end{equation}

\paragraph{Attention Layers} Now a sentence is a series of vectors $(t_0,...,t_{L-1}), ~ t_i\in \R^N$.
Attention is a continuous analogy of a dictionary (hash table) where we need a query $q$, key $k$, and value $v$.
In attention, $q$, $k$ and $v$ are computed from the token by linear transformations denoted by $F_Q,F_K$ and $F_V$
\begin{equation}
    q_i = F_Q(t_i),~k_i = F_K(t_i),~v_i = F_V(t_i).
\end{equation}
The next step is a little complicated; let's focus on the token $t_{L-1}$, and the rest are the same.
We first compute the dot product between $q_{L-1}$ and $k_0,...,k_{L-1}$.
This gives us a series of scalars $(q_{L-1}\cdot k_0,...,q_{L-1}\cdot k_{L-1})$.
Afterward, we treat these inner products as energy and calculate the Boltzmann distribution at a certain temperature.
Finally, the result of applying attention to $t_{L-1}$ is the weighted sum of each $v_i$ according to probabilities given by the Boltzmann distribution.
The above procedures can be formally written as
\begin{equation}\begin{aligned}
    &\textrm{Attention}(q_{L-1};k_0\cdots k_{L-1};v_0\cdots v_{L-1})\\
    &\qquad \qquad\qquad\qquad=\textrm{softmax}(\beta q_{L-1}K^T)V
\end{aligned}\end{equation}
where $K^T=(k_0,...,k_{L-1})$ and the same goes for $V$.
Or put every query together as $Q^T=(q_0,...,q_{L-1})$ and be more concise
\begin{equation}
    \textrm{Attention}(Q;K;V) = \textrm{softmax}(\beta QK^T)V
\end{equation}
After attention, each token, which was a vector in $\R^N$, is mapped to a new vector in $\R^N$.
Finally, a linear layer applying on each token one by one follows, which constitutes an attention layer.
Modern transformers are typically composed of dozens of attention layers stacked together.
The output of the Transformer corresponding to the last token $t_{L-1}$, after another linear layer, is explained as the probability distribution of the next token.
Formally, the computations of Transformer Layers can be written as
\begin{equation}
    \FF\Bigg(\Big(\FF\circ\textrm{Attention}\big(F_Q(t);F_K(t),F_V(t)\big)\Big)\times M\Bigg),
\end{equation}
where $M$ is the number of Transformer layers, $\FF$ stands for a linear layer (with activation functions; different $\FF$ are different linear layers), and $t$ are the token embeddings we got from tokenization and token embedding.

\paragraph{Historical Notes} Before Transformer, people used recurrent neural networks (RNN), especially long short-term memory (LSTM) and gated recurrent neural networks, to process natural language.
Then, the attention mechanism was proposed as an improvement method to the RNN.
Later, in 2017, people found that architecture that relies solely on the attention mechanism works better than RNN with attention, as the quote said \textit{Attention Is All You Need}.
This new architecture is called \textit{Transformer}.

\section{Potts and $O(N)$ models}
\label{sec:reviewOn}

The $N$-state Potts model on the lattice \cite{Potts_1952} is a generalization of the Ising model.
It consists of $N+1$ spin states on each lattice site with nearest-neighbor interaction depending on whether two nearest-neighbor states are different or the same.
The spin states can be represented by $N+1$ vectors $e_i^{\alpha}$, where $\alpha= 1,\cdots, N+1$ and $i=1,\cdots N$ with
\begin{equation}
    \sum_i e_i^{\alpha} e_i^{\beta} = \frac{N+1}{N}\delta^{\alpha \beta} - \frac{1}{N}.
\end{equation}
We associate Greek indices $\sigma,\tau,...$ with lattice points and Latin indices $i,j,...$ with internal (spin) degrees of freedom.
The partition function in the presence of an external source $J$ is then given by (where we are now summing over repeated indices)
\begin{equation}
    Z[J] = \sum_{\{t\}} \exp\left(- \frac{1}{2} t_{i,\sigma} K_{\sigma \tau} t_{i,\tau} + J_{i,\sigma} t_{i,\sigma}\right),
\end{equation}
where $\sigma$, $\tau$ denote lattice sites and the spin variables run over all $N+1$ unit vectors $e_i^{\alpha}$ for the state at each lattice site. $K_{\sigma \tau}$ denotes the coupling matrix and is assumed to be symmetric and vanish unless $\sigma$ and $\tau$ are nearest-neighbors.

Following \cite{Zia_1975}, one can utilize the identity
\begin{align}
    ~ &~ \exp\left(-\frac{1}{2} t_{i,\sigma} K_{\sigma \tau} t_{i,\tau}\right) \nonumber \\
    &= C\left[\prod_{j,\rho}\int d\phi_{j,\rho}\right]\exp\left(\frac{1}{2}\phi_{i,\sigma} (K^{-1})_{\sigma \tau} \phi_{i,\tau} + \phi_{i,\tau} t_{i,\tau}\right),
\end{align}
with $\phi_{i,\sigma}$ being a new variable with $N$ components for each lattice site to obtain a field theory in the continuum limit. To this end, one performs the sum over the spin states $t$ and replaces $\phi_{i,\sigma}$ by a field $\phi_i(x)$. The term $\phi K^{-1} \phi$ can be expanded in local derivatives of the field $\phi$, and after retaining only up to two derivatives and polynomials with up to four powers of $\phi$ (higher order terms can be discarded as irrelevant in the RG flow of the field theory), one obtains the following Euclidean Hamiltonian
\begin{equation}\begin{aligned}
    &\frac{\mathcal{H}}{k_B T} = \int d^d x \left(\frac{1}{2} (\nabla \phi)^2 + \frac{1}{2} r_0 \phi^2 + \frac{1}{3!} q_0 Q_{ijk} \phi_i \phi_j \phi_k\right.\\
    &\qquad\qquad\quad\left. + \frac{1}{4!} (u_0 S_{ijkl} + f_0 F_{ijkl}) \phi_i \phi_j \phi_k \phi_l\right),
\end{aligned}
\end{equation}
with couplings
\begin{align}
    Q_{ijk} &= \sum_{\alpha} e_i^{\alpha} e_j^{\alpha} e_k^{\alpha}, \nonumber \\
    F_{ijkl} &= \sum_{\alpha} e_i^\alpha e_j^\alpha e_k^\alpha e_l^\alpha, \nonumber \\
    S_{ijkl} &= \frac{1}{3}(\delta_{ij}\delta_{kl} + 2\textrm{permutations}).
\end{align}
The renormalization of such theories in higher than four dimensions was initiated in \cite{Amit_1976}. We will be interested in the restricted Potts model with vanishing trilinear coupling, $q_0 = 0$, corresponding to the lattice model with $2(N+1)$ state vectors $\pm e_i^\alpha$. Furthermore, we will specify the most symmetric phase with $O(N)$ invariant quartic interaction and field theory action
\begin{equation}
    S = \int d^d x \left(\frac{1}{2} (\partial \phi_i)^2 + \frac{\lambda}{4}(\phi_i \phi_i)^2\right).
\end{equation}
For $d > 4$ dimensions, the interaction term is irrelevant, and one expects a free theory as an infrared (IR) fixed point. However, the authors of \cite{Fei:2014yja} show the existence of an interacting UV fixed point in $6 - \epsilon$ dimensions. Remarkably, such a fixed point is only unitary for $N > 1038$, giving rise to an interacting conformal field theory (CFT) there. 

In \cite{Mati:2016wjn}, the critical exponents for such UV fixed points in higher dimensions were computed, and it was found that for $N \rightarrow \infty$
\begin{equation}
    \nu = (d - 2)^{-1},
\end{equation}
where $\nu$ is the scaling exponent of the correlation length, i.e. for $|x| \rightarrow \infty$,
\begin{equation}
    \langle t(x) t(0) \rangle \sim \exp(-x/\xi),
\end{equation}
with
\begin{equation} \label{eq:nudef}
    \xi\sim\begin{cases}
        f_+ |T-T_c|^{-\nu}&T>T_c\\
        f_- |T-T_c|^{-\nu'}&T<T_c
    \end{cases}.
\end{equation}
In other words, the correlation length $\xi$ can be viewed as inverse mass and its scaling near the critical point is given by the above law.

\section{Transformer as an $O(N)$ model}
\label{sec:transformerasOn}

The interaction inside the $O(N)$ model is nearest, i.e., only adjacent spinors have interaction.
The interaction in the attention mechanism is non-local.
That is to say, any two tokens have interaction.
However, for a given token, its attention strengths with different tokens are different.
We can pick out the dominant ones and treat other high-order terms as perturbations.
The perturbations will not affect the critical phenomena (``the critical phenomena are rigid").
For example, as the expectation value of the inner product of two random unit vectors $u,v$ in $N$ dimensional space is
\begin{equation}
    \mathbb{E}\left[\left(u\cdot v\right)^2\right] = \frac{1}{N},
\end{equation}
we can use $\frac{1}{\sqrt{N}}$ or $\frac{3}{\sqrt{N}}$ as the threshold for dominant interaction.
In this way, we get a graph in which each site is a token and edges are strong attentions between tokens.
We can embed this graph into a high-dimensional space; in this configuration, the interaction between tokens is the nearest.
The above procedure is demonstrated in Figure \ref{fig:transformerasOn}.
We call the dimension of the high-dimensional space \textit{internal dimension} denoted by $d$.
We conjecture that the internal dimension $d$ is the intrinsic dimension studied in \cite{intrinsicdim2023}.
The internal dimension reflects the complexity of the system, where a higher internal dimension indicates a more complex system.

The energy is defined as
\begin{equation}\label{eq:defE}
    E=\frac{1}{L}\sum_{\sigma=0}^{L-1}\sum_{\tau=0}^{L-1}t_\sigma\cdot t_\tau
\end{equation}
We expect the specific heat near the critical point $T_c$ to have the behavior
\begin{equation}
    C\sim\begin{cases}
        A_+ |T-T_c|^{-\alpha}&T>T_c\\
        A_- |T-T_c|^{-\alpha'}&T<T_c
    \end{cases}
\end{equation}
As a result, if $\alpha,\alpha'<1$, the energy behaves like
\begin{equation}\label{eq:Escaling}
    E\sim\begin{cases}
        E_c + \frac{A_{+}}{1-\alpha}(T-T_c)^{1-\alpha} & T>T_c\\
        E_c - \frac{A_{-}}{1-\alpha'}(T_c-T)^{1-\alpha'} & T<T_c
    \end{cases}
\end{equation}
From \cite{amit1984field}, the exponent $\alpha$ shall have the following relation with the internal dimension $d$
\begin{equation}
    \nu d = 2-\alpha, \quad \nu = \frac{1}{d-2},
\end{equation}
where $\nu$ is the critical exponent of the correlation length \eqref{eq:nudef}.
By measuring the $\alpha$ around $T_c$ we can probe the internal dimension
\begin{equation}\label{eq:alpha_d}
    \alpha(d) = 2 - \frac{d}{d-2} = \frac{d-4}{d-2},\quad d(\alpha)=\frac{2(2-\alpha)}{1-\alpha}.
\end{equation}
Notably, this method of estimating $d$ is independent of the definition at the very beginning of this section.
It does not require specifying which interactions should be ignored and which interactions should be kept.
The dimensionality can be determined solely from the relationship between the energy and temperature. 

\paragraph{RG flow}

As we will see in the Experiments section, the transformer's energy curve fits well with the corresponding curve for the $O(N)$ model in a dimension close to $6$. The fact that the critical exponents come out roughly correctly for this description indicates that the underlying dynamics of machine text generation follow a critical behavior. We conjecture that, 
around the critical temperature, this dynamics is in the same universality class as the $O(N)$ model in higher dimensions at its critical point. 

These findings suggest that the transformer architecture implements a flow from human language to an artificial language. That is, each time the transformer generates a new token, the system has moved one step closer in this direction. After many such tokens have been generated, the system reaches an equilibrium, which we call \textit{machine language}. Since our findings indicate a close connection to the $O(N)$ model at a critical fixed point, we can identify the above flow as an RG (Renormalization Group) flow between two critical systems, the starting point being human language and the endpoint machine language.

\section{Experiments}
\label{sec:experiments}

\begin{figure}[h]
    \centering
    \includegraphics[width=1.0\linewidth]{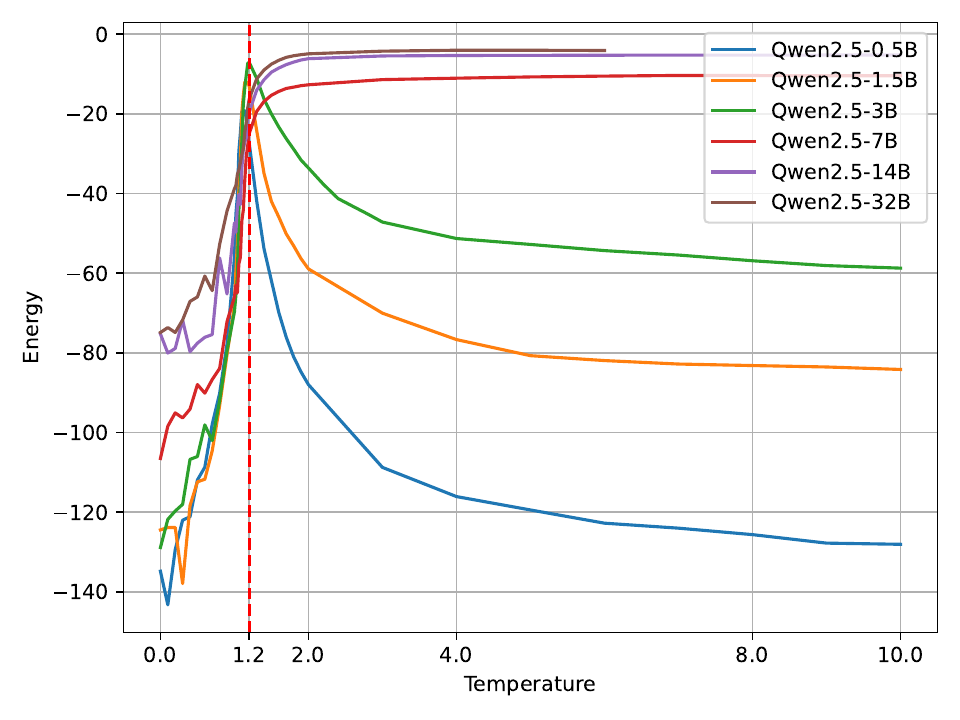}
    \caption{Energy of Qwen2.5 models. The temperature is used in generating text. Higher temperatures result in more random generation. The energy is computed by Eq. \eqref{eq:defE}.}
    \label{fig:energy_qwen_original}
\end{figure}

We tested our theory on the Qwen model series \cite{qwen2.5}.
Qwen offers the most comprehensive range of open-source models with varying parameter sizes, from small to large: 0.5B, 1.5B, 3B, 7B, 14B, and 32B. It also offers several fine-tuned versions: Qwen2.5, Qwen2.5-Math, Qwen2.5-Coder, and Qwen-Coder-Instruct.
We use text from English Wikipedia \cite{simple_wiki} as the seed prompts and let the model generate 1024 tokens at different temperatures $T$ freely.
Then, we compute the average energy per token based on Eq. \eqref{eq:defE} at different $T$.

\begin{figure}[h]
    \centering
    \includegraphics[width=1.0\linewidth]{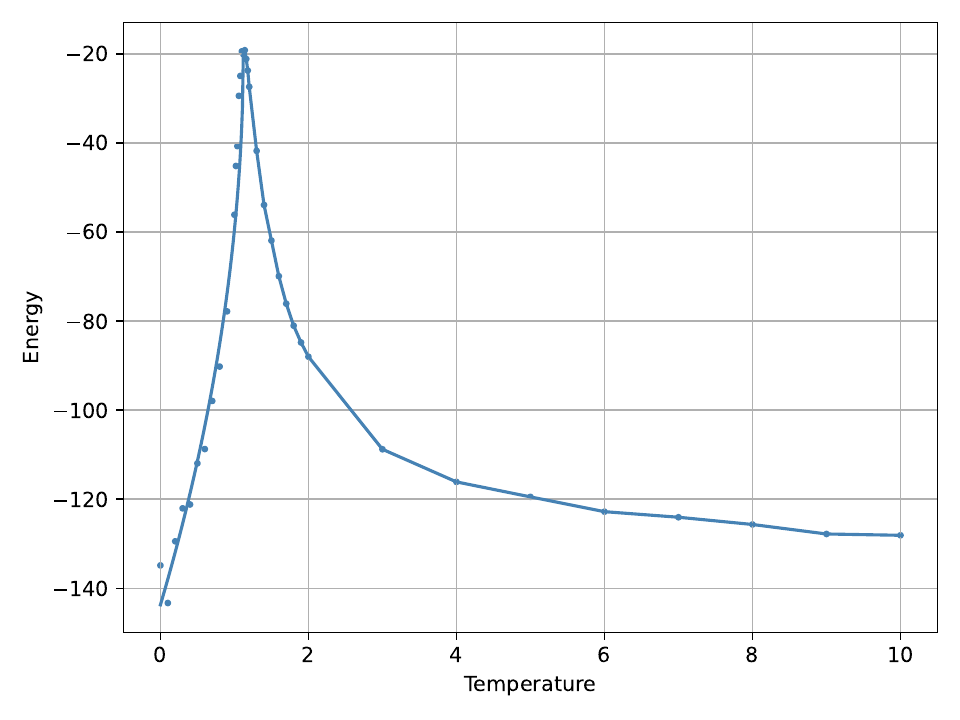}
    \caption{Energy-temperature curve of small LLMs. This figure shows the energy of Qwen2.5-0.5B.}
    \label{fig:qwen0.5b}
\end{figure}

\begin{figure}[h]
    \centering
    \includegraphics[width=1.0\linewidth]{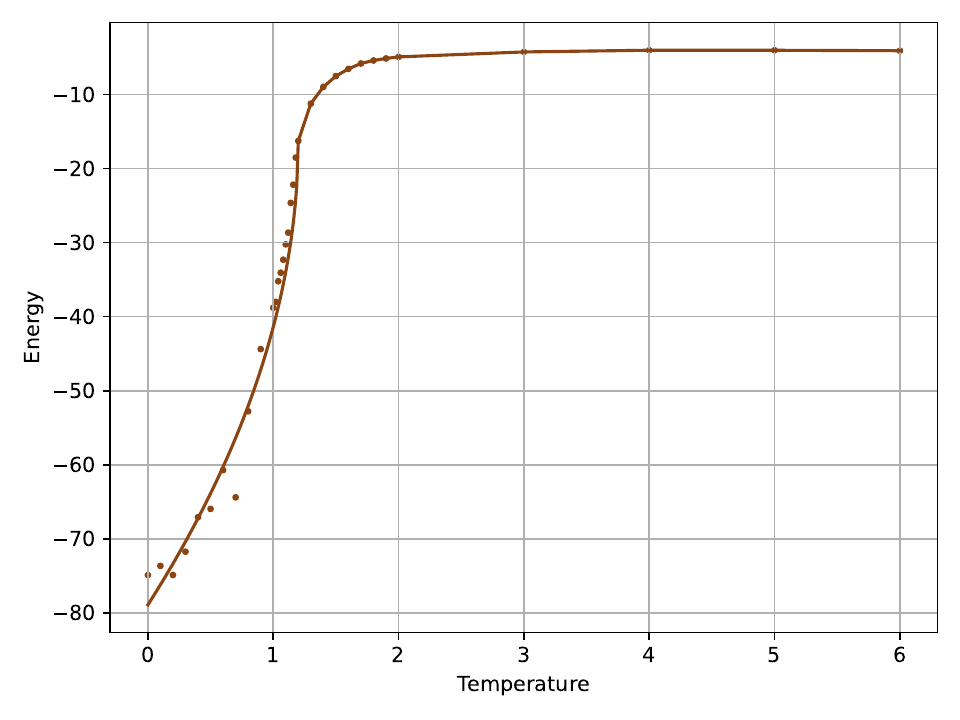}
    \caption{Energy-temperature curve of large LLMs. This figure shows the energy of Qwen2.5-32B.}
    \label{fig:qwen32b}
\end{figure}

\paragraph{Phase transition w.r.t. temperature} As shown in Figure \ref{fig:energy_qwen_original}, the model undergoes a phase transition as the temperature increases.
At lower temperatures, the model's output closely resembles human language (``meaningful phase"), while after the phase transition, the output becomes nonsensical (``nonsense phase").
Surprisingly, models of different sizes exhibit the same phase transition temperature (critical temperature) and the same maximum energy.
The shared critical temperature is approximately $T_c=1.2$, while the maximum energy is around $E_{\textrm{max}}=-4.0$.
Note that, as described in Eq. \eqref{eq:defE}, lower energy indicates stronger interactions between words in a sentence, whereas higher energy corresponds to disordered and random sentences.
Therefore, it is surprising that all models share the same maximum energy.

As the temperature increases from low to high and crosses the critical temperature, the model exhibits second-order phase transition behavior.
A second-order phase transition occurs when the energy remains continuous, as in the case of heating a magnet until it loses its magnetization. 
However, in the nonsense phase, models of different sizes demonstrate distinct behaviors.
Small models (0.5B, 1.5B, 3B) exhibit negative specific heat in the nonsense phase, as shown in Figure \ref{fig:qwen0.5b}.
We will later explain that this phenomenon arises from the small models' inability to recognize that they are generating nonsense. 
In contrast, large models (7B, 14B, 32B) display a behavior similar to the second-order phase transition observed in the Ising model, as shown in Figure \ref{fig:qwen32b}.

\begin{table}[ht]
    \centering
    \begin{tabular}{c|cccccc}
        Model Size & 0.5B & 1.5B & 3B & 7B & 14B & 32B \\
        \hline
        $\alpha'$ & 0.49 & 0.56 & 0.58 & 0.62 & 0.62 & 0.49\\
        $d(\alpha')$ & 5.9 & 6.5 & 6.8 & 7.3 & 7.3 & 5.9\\
        $d_{\textrm{intrinsic}}$ & 6.2 & 5.6 & 5.2 & 5.3 & 5.2 & 5.4
    \end{tabular}
    \caption{The internal dimension of Qwen2.5 series and the intrinsic dimensions of each model at $T=1$.}
    \label{tab:dimqwen}
\end{table}

By fitting Eq. \eqref{eq:Escaling}, we can obtain the critical exponent $\alpha'$ of the model in the meaningful phase.
Using $\alpha'$, we can calculate the spatial dimensionality of the $O(N)$ model corresponding to the Transformer, as described in Eq. \eqref{eq:alpha_d}.
Our results are shown in Table \ref{tab:dimqwen}, where it can be observed that as the model's parameter size increases, the dimensionality of the model's output also increases.
We also list the intrinsic dimensions of each model at $T=1$ in Table \ref{tab:dimqwen}, we can see that the internal dimension $d$ and the intrinsic dimension $d_{\textrm{intrinsic}}$ are of the same magnitude.

\begin{figure}
    \centering
    \includegraphics[width=1.0\linewidth]{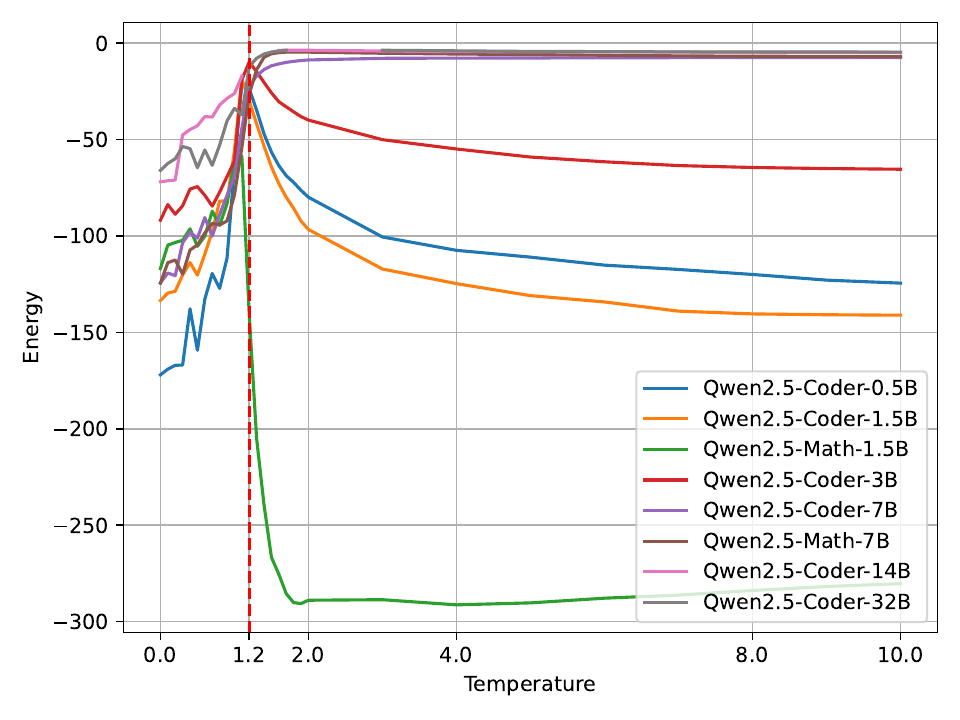}
    \caption{Energy of Qwen2.5-Math and Qwen2.5-Coder models}
    \label{fig:energy_qwen_math}
\end{figure}

We also test the energy curve on the Qwen-Math and Qwen-Coder series.
The result is shown in Figure \ref{fig:energy_qwen_math}.
From these figures, we can see that the phenomenon we just described holds true across various models.

\paragraph{Phase transition w.r.t. parameter size}

In the nonsense phase $T>T_c$, both small and large models produce nonsensical outputs.
However, their awareness of whether they are generating nonsense differs.
The energy $E$ we define measures how much a model perceives a sentence as meaningful.
Lower energy indicates stronger interactions between the tokens that compose the sentence, suggesting the model perceives the sentence as more meaningful.
From the energy plots above, large models are aware that they are generating nonsense, as their interaction energy is very high.
In contrast, small models believe they are producing meaningful outputs, with interaction energy levels almost as low as when $T=0$. 
The ability to recognize that their outputs are nonsensical is a significant emergent behavior \cite{emergent_google_2022}.
In our experiments, only sufficiently large models exhibit this capability.

This is a novel phase transition phenomenon.
When the parameter size is relatively small, the model undergoes a phase transition like Figure \ref{fig:qwen0.5b}.
However, as we continuously increase the parameter size, the transition at \(T_c\) becomes similar to Figure \ref{fig:qwen32b}.
This represents a \textit{phase transition of a phase transition}, which we refer to as a \textit{higher-depth} phase transition.

\begin{figure}
    \centering
    \includegraphics[width=1.0\linewidth]{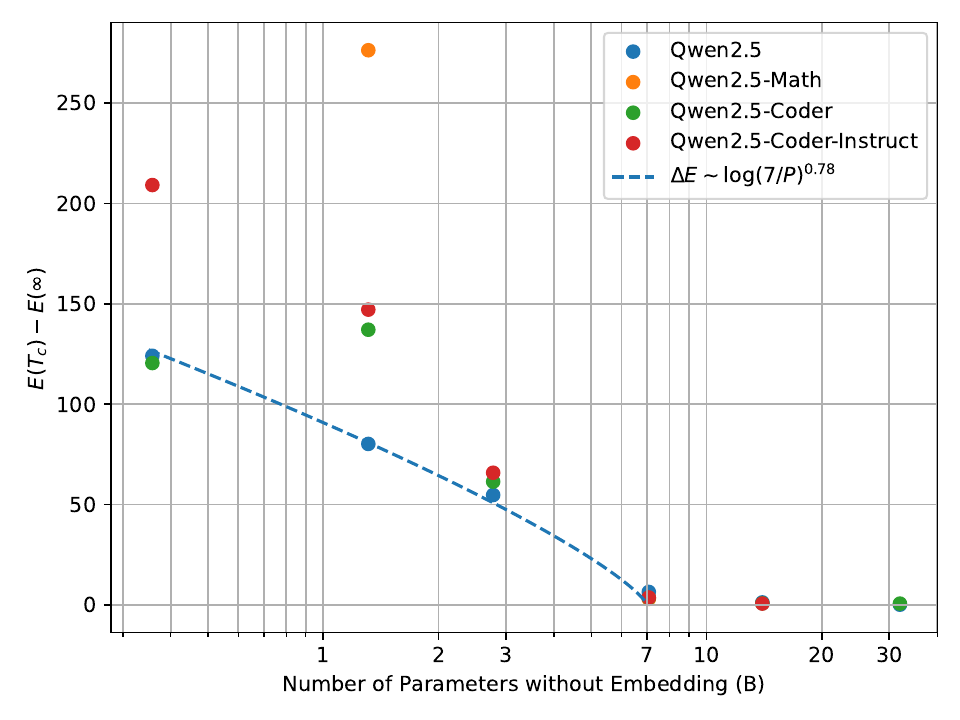}
    \caption{Want to know whether your model's parameter size is sufficient? Measure \(E(T_c) - E(\infty)\).}
    \label{fig:qwen_bloating}
\end{figure}

\(E(T_c) - E(\infty)\) measures whether a model's parameter size is sufficient.
As shown in Figure \ref{fig:qwen_bloating}, the higher-depth phase transition occurs at approximately $P_c=7\textrm{B}$ parameters.
Furthermore, we observe that variants of Qwen, which are trained on more data, exhibit the same phase transition parameter size as the original Qwen2.5.
This suggests that \textbf{small models and large models are fundamentally different entities}.
Despite both being based on the Transformer architecture, large models with more than 7B parameters exhibit entirely different behaviors compared to smaller models with fewer than 7B parameters.


\paragraph{Application}

As an application of our theory, the \(E-T\) curve can be used to determine whether to increase the model's parameter size.
In the training process of LLMs, it is often observed that adding more data does not improve model performance.
At this point, should one focus on improving data quality (e.g., cleaning the dataset) or increasing the model's parameter size? 
Our theory provides a new method to address this question: measure \(E(T)\) and observe whether \(E\) decreases after crossing \(T_c\). If \(E\) decreases, it indicates the need to increase the parameter size. If not, it suggests focusing on cleaning the data. 
This method is efficient (requiring only a few minutes), does not rely on additional data (e.g., a test set), and is highly sensitive.

\section{Related Work}
\label{sec:relatedwork}

LLM scaling laws were first proposed in \cite{scaling_2017, scaling_openai_2020, scaling_rosenfeld_2021}.
From a physics perspective, scaling laws are almost synonymous with phase transitions, critical phenomena, quantum field theory, fractals, and percolation theory.

\paragraph{Fractals}
\cite{intrinsicdimensiondata_2019} studies data representations' intrinsic dimension (fractal dimension) in deep neural networks.
\cite{intrinsicdimensionobjective_2018} measure the intrinsic dimension of the neural
network in parameter space.
\cite{scalingintrinsicD2022} relates the fractal dimension of the data and the scaling laws. They imagine the data used in training neural networks form a fractal. Denote the dimension of the fractal by $d$; they argued that the parameter scaling exponent is related to $d$ as $\alpha=4/d$.
\cite{intrinsicdim2023} shows an example of using the fractal dimension in the detection of AI-generated texts.

\paragraph{Quantum Field Theory}
\cite{solvablemodelneuralscaling_2022} proposed a statistical model and solved it to explain the neural scaling phenomenology.
\cite{neuralscalinglawslargen2024} solved the previous model using large-$N$ field theory methods.
\cite{Halverson_2020, Halverson_2021, Halverson_2024} describe how to construct Quantum Field Theory (QFT) from neural networks.
\cite{Giataganas_2022} found evidence of the connection between neural networks and the Renormalization Group (RG) flow.

\paragraph{Percolation Theory}
\cite{percolation2022} explores the resilience of neuronal networks to damage using inverse percolation and identifies a phase transition in network connectivity.
\cite{percolation2024} establishes a phase transition model by drawing an analogy between neural network learning dynamics and percolation on a bipartite graph, providing a theoretical framework to predict and understand the emergence phenomena observed in neural networks. They also argued that emergence is different from grokking.

\section{Conclusion}
\label{sec:conclusion}

In this paper, we have analyzed the transformer architecture from the point of view of critical phenomena.
Inspired by similarities to the $O(N)$ model in physics, each token can be viewed as an internal spin with Attention facilitating spin-spin interactions.
We measure the \textit{energy} of generated text and plot it as a function of temperature.
We find that these curves display a phase transition with a critical temperature near which the energy follows a scaling law.
From the scaling exponents, we can extract the internal dimension of the $O(N)$ model.
We interpret this as an RG flow from natural language to machine language, with each new token constituting a flow step.
Two distinct behaviors are found for small and large models.
The energy of small models is small for $T > T_c$, while the energy of large models is relatively much larger for this temperature range.
This indicates that large models are \textit{aware} that they are producing nonsense text, while smaller models are not.
We interpret this as an emergent capability.

\section*{Acknowledgements}

We would like to thank Serguei Barannikov and Cui Wei for their valuable discussions.
The work of B.H. and Y.S. is supported by NSFC grant 12250610187.

\section*{Impact Statement}

This paper presents work whose goal is to advance the field of 
Machine Learning. There are many potential societal consequences 
of our work, none of which we feel must be specifically highlighted here.

\nocite{langley00}

\bibliography{example_paper}
\bibliographystyle{icml2025}

\newpage
\appendix
\onecolumn



\end{document}